\documentclass[sigconf]{acmart}

\usepackage{booktabs} 
\usepackage{adjustbox}
\usepackage{tabularx}
\usepackage{url}

\usepackage[ruled]{algorithm2e} 

\SetAlFnt{\small}
\SetAlCapFnt{\small}
\SetAlCapNameFnt{\small}
\SetAlCapHSkip{0pt}
\IncMargin{-\parindent}

\setcopyright{rightsretained}

\acmDOI{10.475/123_4}

\acmISBN{123-4567-24-567/08/06}

\acmConference[WOODSTOCK'97]{ACM Woodstock conference}{July 1997}{El
  Paso, Texas USA}
\acmYear{1997}
\copyrightyear{2016}

\acmArticle{4}
\acmPrice{15.00}

\editor{Jennifer B. Sartor}
\editor{Theo D'Hondt}
\editor{Wolfgang De Meuter}

\begin{document}
\title{Mining Procedures from Technical Support Documents}

\author{Abhirut Gupta}
\affiliation{
\institution{IBM Research AI}
}
\email{abhirutgupta@in.ibm.com}
\author{Abhay Khosla}
\affiliation{
\institution{IBM Research AI}
}
\email{khoslaab@in.ibm.com}
\author{Gautam Singh}
\affiliation{
\institution{IBM Research AI}
}
\email{gautamsi@in.ibm.com}
\author{Gargi Dasgupta}
\affiliation{
\institution{IBM Research AI}
}
\email{gaargidasgupta@in.ibm.com}


\renewcommand{\shortauthors}{}

\begin{abstract}
Guided troubleshooting is an inherent task in the domain of technical support services. When a customer experiences an issue with the functioning of a technical service or a product, an expert user helps guide the customer through a set of steps comprising a troubleshooting procedure.  The objective is to identify the source of the problem through a set of diagnostic steps and observations, and arrive at a resolution. Procedures containing these set of diagnostic steps and observations in response to different problems are common artifacts in the body of technical support documentation. The ability to use  machine learning and linguistics to understand and leverage these procedures for applications like \textit{intelligent chatbots} or \textit{robotic process automation}, is crucial.  Existing research on question answering or intelligent chatbots does not look within procedures or deep-understand them. In this paper, we outline a system for mining procedures from technical support documents. We create models for solving important subproblems like  extraction of procedures, identifying decision points within procedures, identifying blocks of instructions corresponding to these decision points and mapping instructions within a decision block. We also release a dataset containing our manual annotations on publicly available support documents, to promote further research on the problem.
\end{abstract}

%
%



\maketitle

\section{Introduction}
Technical support refers to a variety of after sale services provided to the customer by the manufacturer of a technology product. A low barrier for entry in the technology sector, specially in software, combined with booming consumer markets across the world, has led to innumerable products offered from small and medium scale companies. Services arms of large organizations also maintain and support a wide variety of products and their versions for enormous customer bases. In such a scenario, traditional models of support provided by human experts, just do not scale, and automation of support is the solution.

\begin{figure*}
\begin{center}
\includegraphics[width=0.8\textwidth]{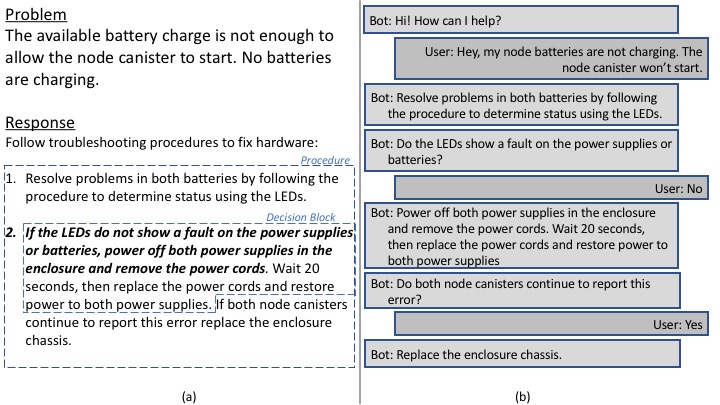}
\end{center}
\caption{(a) Snippet of a procedure and context in a support document on the web, with the procedure identified and the decision block extracted (b) Sample conversation by an automated bot using the procedure}
\label{procexample}
\end{figure*}

Research in the domain of technical support has tackled various important problems like complex question understanding, knowledge graph based question answering, and conversational agents~\cite{dhoolia2017cognitive}. However, one important problem that has not garnered significant attention is the deep understanding procedures in technical support documents. Procedure seeking ``How to" questions form the second largest class of questions in web queries~\cite{de2005question}. Additionally analysis of support tickets reveals that about 26\% of user problems are ``How-to"s, directly seeking a procedure (refer Table~\ref{distrib_question}). Problems requiring complex troubleshooting form around 48\% of support questions and also stand to benefit from understanding procedures in content. Diagnosis of user problems and subsequent troubleshooting needs to be performed by following a set of steps described within procedures. Figure~\ref{procexample}(a) shows an example procedure in the hardware domain which details the troubleshooting response to the problem ``node cannisters won't start''. 
Common tasks like part replacements or maintenance have standard procedures associated with them.  The product documentation is a treasure trove of this information, containing a large number of procedures for diagnostics, troubleshooting, maintenance, or common tasks. From our manual annotations, we find that about 58\% of support documents across three IBM products contain at-least one procedure. However, these procedures are present in the document as natural language text, and are not ``understood" by systems answering technical support questions. State of the art approaches today aim at returning relevant support documents to a user problem~\cite{yang2017efficiently}, and these documents usually contain procedures. However the internals of the procedures are not identified by the algorithms.

\begin{table*}
  \caption{Distribution of 202 support questions across various IBM software products}
  \label{tab:commands}
  \begin{tabular*}{\textwidth}{ccp{0.6\textwidth}}
    \toprule
    Question class&Number of questions&Example\\
    \midrule
     Require Troubleshooting& 97 (48\%) &Hi Team, GCE site performance is too slow in both Staging and Production site after Version upgrade. Kindly check. Thank you.\\
     How-To& 53 (26\%) &We are on Cognos 10.2.2 on Linux servers.  We need instructions on how to load certificate on our webserver where we are storing graphics.\\
     Require Information (Factoid)& 39 (19\%) & We need to upgrade Netezza version to 7.1.0.4 - P2 and Latest Firmware. Let us know available slots for the upgrade.\\
     Action Request & 13 (6\%) &Hi Team,    Please reactivate the below SPSS v23 concurrent codes <code1> <code2> <code3>\\
    \bottomrule
  \end{tabular*}
  \label{distrib_question}
\end{table*}

Conversational chat-bots and Robotic Process Automation (RPA) are two important applications that can benefit from mining procedures. In the world of RPA, developers refer to procedure documents supplied by business teams for their automation script building. Today the business teams create these documents manually which is both time and labor consuming. A more scalable approach would be to mine these from the existing repository of product documentation and technical forums. 
Automated procedure extraction from technical documents helps reduce automation documentation creation time.
Similarly conversational chat-bots could auto-bootstrap their dialog flows from these extracted procedures. Decision points and the corresponding blocks within procedures could be presented as choices to the user. This enables quick creation of intelligent chat-bots for guided troubleshooting without much manual effort.
The value statement of mining procedures is large especially because it leverages already existent documentation. 
In contrast, another approach could be to manually create structured procedure flows for these applications. Apart from being time consuming, a possible drawback is that the structure would then differ across applications and possibly become vendor specific. We also argue that it is much easier and intuitive for humans to create natural language descriptions for procedures than these structured application specific flows. Hence, technical manuals/product documents written in natural language forms the input to the procedure mining task. 

Figure~\ref{procexample}(a) shows a support page we annotate, and (b) shows a sample conversation by a chat-bot using this extracted procedure. In Figure~\ref{procexample}(a) we also highlight the section of the page identified as the procedure, the sentence in bold as an example of a ``decision point'', and the set of sentences following the condition in the decision point labeled as the ``decision block''. A procedure may contain one or more decision points and corresponding decision blocks. 

For the purpose of our investigation, we define procedures as a sequence of steps which need to be executed (possibly in a non-linear flow) to achieve a target goal. 
The goal in technical support is often `diagnosing a problem', `troubleshooting/resolving a problem', or `replacing a part' but we do not restrict the mining task to these fixed set of goals. In this paper, we address the problem of mining technical procedures from support documents written in natural language. 
Our contributions are as follows:
\begin{enumerate}
\item We create a system to (a) identify and (b) extract procedural text from support documents
\item We work on identification and extraction of decision points within procedure steps, corresponding decision blocks, and mapping of instructions within decision blocks to appropriate paths.
\item We release a dataset, crawled and manually annotated from the publicly available website, Knowledge Center, to promote further research in the area.
\end{enumerate}

\textbf{Data:} For each of the tasks identified - procedure identification and extraction, decision block extraction, and mapping of instructions within decision blocks, we collect data by annotating technical support documents for three IBM products - IBM Campaign, Digital Analytics and Storwize. In all, we annotate data from 532 pages. For procedure identification, we annotate lists as positive and negative examples of procedures. About 949 lists are annotated, from which we find 405 to actually be procedures. For decision block extraction, we annotate 237 instances of text following a decision point. In each of these blocks we also map sentences to appropriate paths at the split created due to the decision point.

\section{Related Work}
Previous research on procedural text falls broadly in two categories. The first area of research deals with extracting a Knowledge Base for procedural text, citing gaps in conventional Knowledge Graphs and their inadequacy in answering "how-to" questions. \cite{chu2017distilling} aims to extract tasks and their parameters from task headers in WikiHow database and organize them into a hierarchical Knowledge Base. \cite{park2018learning} proposes a similar aim, to recognize actions in procedures, and organize them in a Knowledge Graph with procedure specific relations. In \cite{Yang:2015:LPK:2766462.2767744}, the authors align queries with tasks in procedural text to improve ``task suggestion'' and for procedural knowledge base creation. While all these methods aim at improving Information Retrieval for ``how-to'' web queries, they do not aim to extract whole procedures for downstream tasks like chatbot, or automation.

The second category of work focuses on extraction of complete procedures from text. 
In \cite{delpech2008investigating} the authors present a thorough linguistic categorization of procedural text, and aim to identify procedure titles, instructions, and simple instructional compounds. \cite{zhang2012automatically} proposes a representation for extracted procedures, and uses rules on top of syntactic parses to extract elements of procedures. \cite{bandyopadhyay2012question} aim to perform question answering by organizing procedural text into instructional compounds. While this body of work is similar to ours in spirit, all these approaches assume the presence of procedural text. Their approaches also fall short of handling conditional sentences, which form decision points in procedures, and extracting complex decision blocks.

To the best of our knowledge, our system is the first to handle identification and extraction of procedures from a corpus of support documents on the web. We differentiate this approach from related work by focusing on identification and extraction of decision points or conditionals in procedural text. These decision points play a major role in support procedures in the form of diagnostic tests. We also bring to focus the problem of extracting decision blocks that possibly span multiple instructions and mapping of instructions inside these blocks to the appropriate path.

\section{Identification of Procedures}
The first step towards mining procedures in technical support documents is to correctly identify them in text. While a significant percentage of a typical support corpus has procedures, a lot of it is also product specification or documentation. From our manual annotation of 532 support pages across three products - IBM Campaign, Digital Analytics, and Storwize, we find procedures in 309 of them. Second, it is important to exactly locate the span of the procedure within a document and separate it from other related information which might also be present in the page. Consider the content of a webpage in Figure~\ref{procexample}(a), after the webpage has been cleaned for templates, hyperlinks and other clutter (not depicted in the figure), the problem still remains to separate the text span of the procedure from text describing the problem and introducing the procedure.

Technical and product documentation comes in various shapes and forms; however, web pages are the most prevalent format today. Most tech companies now provide their diagnostic and troubleshooting procedures, along with FAQs and procedures for common tasks on the web~\footnote{\url{https://support.microsoft.com/en-in}, \url{https://support.lenovo.com/in/en/}, \url{https://www.ibm.com/support/knowledgecenter/}, \url{https://www.ibm.com/support/home/}, \url{http://www.dell.com/support/contents/in/en/inbsd1/category/Product-Support/Self-support-Knowledgebase}}. While we do encounter the occasional PDFs and Word documents, they form a very small percentage of the overall content. These formats can also be easily converted to more consumable HTML markup pages with open tools. Therefore, we restrict the scope of our investigation to HTML web pages. Further, the markup for lists (ordered and unordered) in web pages is an obvious way to represent procedures which are essentially trees (or in more complex cases, graphs) of instructions. In our broad experience with support documents on the web from multiple organizations and products, we only come across a handful of pages where technical procedures are not expressed as lists. Our baseline model for identification of procedures works with only the text content of candidate chunks, but we show that markup information from lists can substantially improve accuracy.

\begin{figure}
\includegraphics[width=\linewidth]{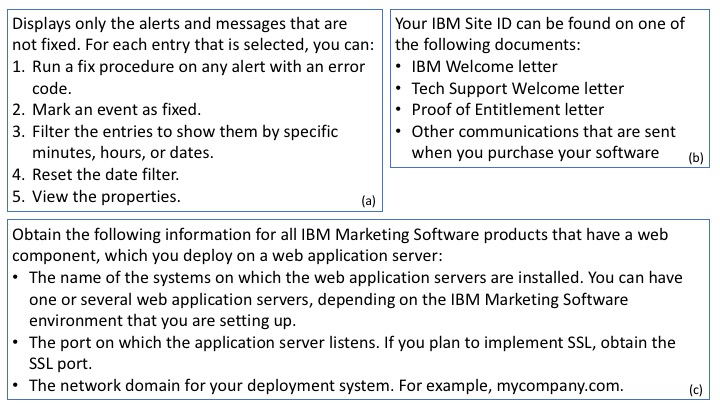}
\caption{Lists identified as negative examples of procedures}
\label{ListNegExamples}
\end{figure}
Given these assumptions, we define the problem of identifying procedures as that of classifying lists on technical support web pages. The discriminative features for identification of procedural lists lie in (a) the words within the text, (b) the contextual sentences around it, (c) formatting of these pages and (d) finally the linguistic nature of the sentences within the list.

To facilitate this classification, we annotate 949 lists from the earlier mentioned 532 pages, and find that 405 lists are actually procedures. Figure~\ref{ListNegExamples} presents a few examples of lists we annotated, which are not considered procedures, as these lists do not describe an order on actions to be performed to achieve a goal. While it might be easier to rule out lists of `items' in Figure~\ref{ListNegExamples}(b), the other two cases are complex. The list of possible actions in Figure~\ref{ListNegExamples}(a) is in fact very difficult to distinguish from a procedure as it uses imperative sentences which are a common feature of procedures. After scraping the content of the pages and aligning it with the annotations, we scrub the page of headers and other \textit{template} information. This is done as our annotations are acquired from only three products and a single website. Features learnt from these templates for identifying procedures might not generalize well to content from other products or pages. All of this information is fed into our feature extractor. All models, exploring the different feature sets, are trained and tuned on around 759 lists in our training set with cross-validation. Results are reported by applying these tuned models to the held out test set of about 190 lists. We use an SVM with a polynomial kernel and L2 regularization for classification.
\subsection{Features}
The first set of features extracted from lists includes the whole text of the list and the text of the context. Later, we also use formatting information from lists for classification.
\subsubsection{Text of the list} As mentioned earlier, template specific information like headers are scrubbed away. These can bias the classifier for the particular set of documents in our dataset but might not generalize to new sources of support documents. To further remove words specific to this dataset, words from context and list text that do not appear in the top 10,000 most frequent words in the Google n-gram corpus are filtered out. We compare the performance of our models with tf-idf bag of word features using upto n-grams with varying n. Test accuracy for models with different n-gram features is presented in Table~\ref{tab:ngram_range}. We find that the unigram model performs best, which might be because of the large number of features generated in the model with bigrams and trigrams compared to the number of training examples. 
\subsubsection{Context} By ``context'' we refer to the sentences introducing the procedure and possibly its purpose. Our intuition is that sentences which introduce a list are a good signal for deciding if it is a procedure, with phrases like \textit{the following steps to <X>} being glaring clues. However, it's not clear how much text before the list should be included before it starts hurting performance. Table~\ref{tab:context_size} compares the results of models by varying the number of preceding sentences used in the context. We find that taking a single sentence before the list gives the best accuracy. Sentence tokenization on text scraped from HTMLs is not highly accurate, however.

\begin{table}
  \caption{Average accuracy over different n-gram ranges}
  \label{tab:ngram_range}
  \begin{tabular}{cc}
    \toprule
    Upto n-gram & Average Accuracy (\%)\\
    \midrule
     Unigram & \textbf{86.0}\\
     Bigram & 84.72\\
     Trigram & 83.83\\
    \bottomrule
  \end{tabular}
\end{table}

\begin{table}
  \caption{Average accuracy over different context sizes}
  \label{tab:context_size}
  \begin{tabular}{cc}
    \toprule
    Context Size (number of sentences) & Average Accuracy (\%)\\
    \midrule
     1& \textbf{86.0}\\
     2& 84.82\\
     3& 83.73\\
     4& 83.92\\
    \bottomrule
  \end{tabular}
\end{table}

\begin{figure}
\includegraphics[width=0.8\linewidth]{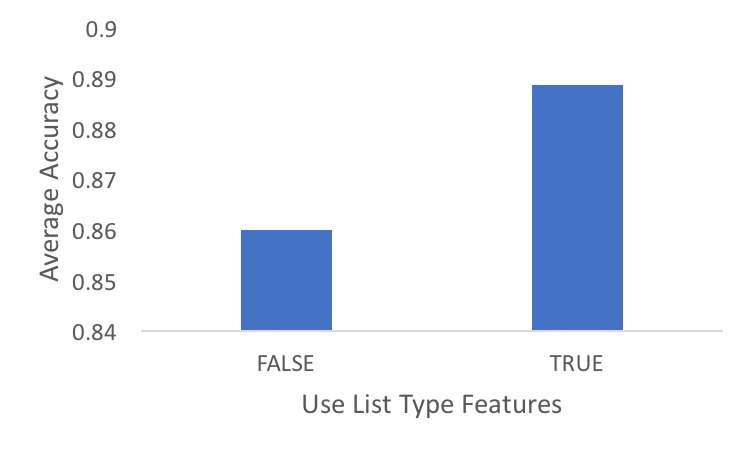}
\caption{Average accuracy improvement by using list-type feature}
\label{list_type_acc}
\end{figure}

\subsubsection{Formatting}
Our next feature investigation includes the list type used for presenting the procedure. HTML lists come in two flavors - ordered and unordered list. Ordered lists define a fixed sequence among the list elements. Given that a procedure is a sequence of steps, ordered lists are a more appropriate method of representing procedures in web-pages. We confirm this hypothesis by including the type of list as a binary feature in our model. Figure~\ref{list_type_acc} shows the substantial improvement in accuracy we achieve from including the list type feature.

In the next sections we discuss the linguistic features we incorporate in the identification algorithm. We use these features for identification of instructions in natural language text. Procedures, which are essentially sequences of instructions often contain dense occurrences of such features.

\subsection{Imperatives} Imperatives are verbs that convey instructions or commands in a sentence. The sentence ``Wait for 30 seconds'' begins with an imperative verb \textit{wait}. Since procedures are essentially a list of steps to achieve a goal, they make heavy use of such imperatives. Imperative words are identified by parsing the given sentence using a specialized parser that returns the mood of every verb used along with other details about each word. Grammatically speaking, every verb used in a sentence in the English language could have one of the following 4 moods - \emph{Imperative} as in ``Restart the PC'', \emph{Indicative} as in ``You restarted the PC'', \emph{Subjunctive} as in ``If you were to restart the PC ...'' or \emph{Infinitive} as in ``To restart the PC, press the button''. In this work, we make use of slot-grammar parser~\cite{DBLP:journals/ijait/McCordBLZ92} to retrieve such word senses.

Slot-grammar is a general framework that offers a way of writing sentence parsing rules for many natural languages. In this paper, we use the slot-grammar parser that was developed by trained linguists for the English language. Broadly, building a grammar for a language using this framework involves writing word-frames for every word in the vocabulary. These word-frames specify rules as to how that word associates syntactically and semantically to other words in a sentence. The parsing algorithm uses these grammar rules to generate multiple parse-trees of a given sentence and assigns a parse score reflecting goodness of each parse-tree hypothesis. We typically make use of the highest scoring parse tree. But in certain cases when the highest scoring tree is likely to be mistaken, say when we know that \emph{drive} is almost always a noun rather than a verb in any technical support document, we may also make use of the second-best or lower scoring parse-tree hypotheses.

One of the conveniences of using such a rule-based parser is the developer's ability to customize it. Technical support documents typically can contain unusual vocabulary and usages and therefore a good parser should be cognizant of them. For example, ``FTP the file'' has the word \emph{FTP}\footnote{short for File Transfer Protocol, a common usage in technical support terminology} used in an imperative verb sense. Furthermore, a sentence like ``Replace the LAN to LAN cable'' usually stumps general-purpose parsers that are unaware that \emph{LAN to LAN cable} is a single noun entity. The English slot-grammar (ESG) parser allows the developer to add new domain-specific lexicon and specify its usage-style.

Through an analysis of this imperative identification method, we demonstrate its reliability as a subtask of procedure identification. We take a random set of 200 sentences from the procedure documents with human labels of imperative verbs. We found both our precision and recall to be 0.82. False negatives and false positives are mostly due to mistaken parse-trees of certain complex sentences. 

Presence of imperatives in text is an important indicator of a procedure. We create features from extracted imperatives like the number of sentences or steps with atleast one imperative, the number of sentences or steps that start with an imperative, and the density of extracted imperatives in text. These features are appended to the already extracted bag-of-word features. We find that extracting imperatives from text improves the accuracy of identification. Although the list type is an essential feature, we might not always find it present. For example, in HTMLs converted from PDFs or Word documents, a procedure might be extracted as a list of paragraphs, or the list type might not be accurate. We find that imperative features help substantially in these cases. Table~\ref{imp_features} presents the improvements in accuracy from imperative features both in cases when list type features are used and when they're omitted. We achieve an accuracy of around 90\% when both list type and imperative features are used.

\begin{table*}
  \caption{Accuracy improvement from using imperative features}
  \begin{tabular}{ccc}
    \toprule
    List type feature used & Imperative features used & Average Accuracy (\%)\\
    \midrule
     No& No& 86.0\\
     No& Yes&\textbf{87.37}\\
     Yes& No&88.86\\
     Yes& Yes&\textbf{90.04}\\
    \bottomrule
  \end{tabular}
  \label{imp_features}
\end{table*}

\subsection{Search Algorithm}
With the procedure identification classifier in place, we leverage the tree structure of HTML documents to do a breadth-first traversal. At each internal node which is a list (ordered or un-ordered), we invoke the classifier to identify if the list is a procedure. We further traverse the children of this node only if it is classified as not a procedure by the classifier. This top-down search for lists is required for two reasons. The first is that lists are often used as tools for formatting the look of a webpage. So while a high level list might be used to organize contents of a webpage, a sub element containing a list might still contain a procedure. On the other hand, a list describing a procedure might often contain sublists that are not procedures but list of items required at some step. An optional thresholding of classification score can be used if the model returns a probabilistic confidence. The algorithm for this search is formalized in Algorithm~\ref{alg:one}.

\begin{algorithm}[t]
\SetAlgoNoLine
\KwIn{Root node ($root$) of the parsed HTML DOM. Optional: Threshold ($t$) of confidence in procedure classification.}
\KwOut{A list of procedures identified in the HTML ($procedures$)}
$procedures = []$; $to\_traverse = [ root ]$;\\
\Repeat{$len(to\_traverse) > 0$}{
        $node = to\_traverse.pop()$\;
        \eIf{
        	$node.label() == ``ol"$ \text{or} $node.label() == ``ul"$\\
        }{
        	$is\_procedure, confidence = classify(node.context(), node.text(), node.label())$\;
            \eIf{
            	$is\_procedure == True$ \text{and} $confidence >= t$\\
            }{
            	$procedures.append(node)$\;
            }{
            	$to\_traverse.add\_all(node.children())$\;
            }
        }{
        	$to\_traverse.add\_all(node.children())$\;
        }
      }
\caption{Procedure Search in Webpage}
\label{alg:one}
\end{algorithm}


\section{Identification of Decision Points}
While procedures for simple tasks like a minor part replacement might be a linear list of steps, most procedures for more complex tasks like troubleshooting involve ``decision points''. As an example, consider the sentence \textit{If the LEDs do not show a fault on the power supplies or batteries, power off both power supplies in the enclosure and remove the power cords} in Figure~\ref{procexample}(a). The flow of instructions in a procedure bifurcates as the result of such decision points, and a particular instance of the procedure would only follow one of the alternate paths created by such a split. It is therefore crucial to identify these decision points in order to correctly mine these procedures for use in downstream tasks like automated chat-bots.

Decision points manifest in text as \textit{conditional sentences} which consist of a \textit{condition} based on which the procedure flow splits, and a \textit{dependent clause} or \textit{effect} which is the instruction to be executed on one particular branch of the flow. In Table~\ref{example_conditional} we show a few examples of conditional sentences with the condition and effect extracted by our system. Most standard kind of sentences that contain condition and effect are those that make use of \emph{if} and \emph{when}.
\begin{itemize}
\item \emph{If the light is blinking}, replace the battery.
\item \emph{When the light is blinking}, replace the battery.
\end{itemize}
In some cases we expect some simple juxtaposition of condition and effect as in 
\begin{itemize}
\item Replace the battery, \emph{if the light is blinking.}
\item Replace the battery, \emph{when the light is blinking.}
\end{itemize}
Some sentences may also be expected to split the \emph{effect} into two parts as in
\begin{itemize}
\item \emph{Replace the battery} when the light is blinking, \emph{or call the technician.}
\end{itemize}
Capturing the condition phrase from within a sentence, in each such sentence type, is the primary step in the task of decision point identification. The remaining portion of the sentence is then deemed as the \emph{effect} part of the sentence.

As is the case with imperative verbs, capturing condition phrases requires identifying the word sense in which the words \emph{if} and \emph{when} are used. This is necessary as not all usages of these words imply a decision point. For example,
\begin{itemize}
\item Check \emph{if} the light is blinking.
\item \emph{When} can a technician be called?
\end{itemize}
The first sentence is not a decision point in itself but is usually followed by a decision point like ``If not''. The second is plain a question and not a decision point.

\begin{figure}
\begin{center}
\includegraphics[width=0.5\linewidth]{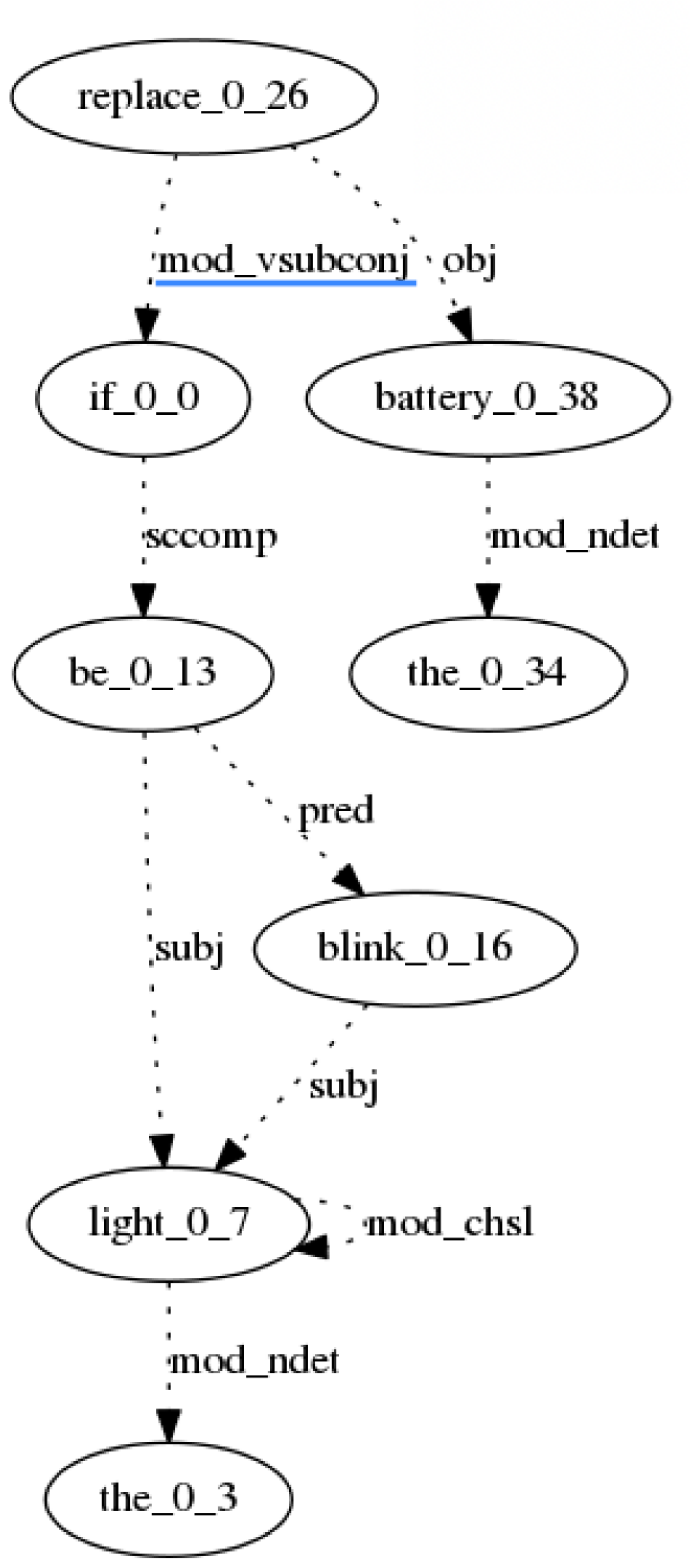}
\end{center}
\caption{Parse-tree of the conditional sentence \emph{If the light is blinking, replace the battery.} }
\label{fig:conditionaltreeexample}
\end{figure}

Hence, we make use of the ESG parse-tree to identify the word sense of the words like \emph{if} and \emph{when}. If the word sense is a \emph{subordinating conjunction}\footnote{A conjunction between a main clause and a subordinate or dependent clause.} that emanates out of a main verb-clause, then we say that a condition is identified. An illustration of this is shown in figure \ref{fig:conditionaltreeexample} which shows an edge labeled \emph{mod\_vsubconj} which stands for \emph{verb subordinating conjunction}. The word senses for every word have been omitted in the figure for brevity. To construct the text snippet of the condition, we use the nodes or words that lie in the subtree rooted at the \emph{if} or the \emph{when} node. The remaining nodes in the tree form the effect snippet. With manual annotation on 215 sentences, we find that the precision of conditional identification is 96\% and recall is around 86\%. Accuracy of condition and effect extraction is found to be around 78\%. Again, most errors are due to incorrect parse trees in complex sentences or slight grammatical or structural mistakes.

\begin{table*}
  \caption{Examples of decision points (\textit{conditional sentences}), with the \textit{condition} and \textit{effect} identified}
  \label{example_conditional}
  \begin{tabular*}{\textwidth}{p{0.3\textwidth}p{0.3\textwidth}p{0.3\textwidth}}
    \toprule
    Conditional Sentence &Condition& Effect\\
    \midrule
     Unless both nodes in the I/O group are online, fix the problem that is causing the node to be offline first& both nodes in the I/O group are online& fix the problem that is causing the node to be offline first \\
     If the LEDs do not show a fault on the power supplies or batteries, power off both power supplies in the enclosure and remove the power cords& LEDs do not show a fault on the power supplies or batteries& power off both power supplies in the enclosure and remove the power cords\\
     When you have performed all of the actions that you intend to perform, mark the error as ``fixed"& you have performed all of the actions that you intend to perform & mark the error as ``fixed"\\
     Swap the drive for the correct one but shut down the node first if booted yes is shown for that drive in boot drive view& booted yes is shown for that drive in boot drive view& shut down the node first\\
    \bottomrule
  \end{tabular*}
\end{table*}

\subsection{Extraction of Decision Blocks}
Once decision points in the procedure have been identified, we focus our attention to ``decision blocks''. We define a decision block as the set of instructions immediately following a decision point, which would be considered only on a single path of execution. Consider the decision block annotated in Figure~\ref{procexample}(a). The instruction \textit{Wait 20 seconds, then replace the power cords and restore power to both power supplies} is only applicable if the condition \textit{If the LEDs do not show a fault on the power supplies or batteries} evaluates to true. The following instruction - \textit{If both node canisters continue to report this error replace the enclosure chassis}, however, applies to both results of the original condition and is therefore not a part of the decision block. Figure~\ref{decision_block_flow_chart} illustrates this flow of instructions. To study this complex problem in detail, we get manual annotations for 237 decision points in our dataset. We ask the annotators to identify sentences directly following a decision point which would be applicable on exactly one value of the condition clause in the decision point. We find that 125 decision points have atleast one following sentence in the decision block. The average number of sentences in a block is 2.36 and the maximum is 15. Although, as mentioned before, sentence tokenization in text scraped from HTMLs is not always exact.
\begin{table}
  \caption{Accuracy with different models for decision block extraction}
  \label{block_accuracy}
  \begin{tabular}{p{0.6\linewidth}c}
    \toprule
     Model & Accuracy (\%)\\
    \midrule
     Baseline & 78\\
     +Removed Note and Information& 84\\
     +Next step conditional overlap& 86\\
     +Stop at sub-list item or paragraph& 90\\
    \bottomrule
  \end{tabular}
\end{table}
Despite the linguistic challenges of identifying the decision block, we find that a simple baseline of considering all sentences following the decision point till the end of the current ``step'' gets about 78\% of our decision blocks correctly. This highlights the importance of structural information about the procedure in its accurate extraction. Following are the major classes of mistakes made by our simple baseline. We try to capture most of them with rules on top of the baseline. Table~\ref{block_accuracy} presents a summary of the accuracy improvement from each of them.
\begin{itemize}
\item Our annotators uniformly reject the presence of \textit{information} sentences such as those starting with ``Note" in decision blocks, as these are not instructions and are considered to be applicable to both branches of a decision point. When present in the same step as a decision point, we apply a simple rule to stop before information sentences, identified using keywords like ``Note'', or ``Information''. Capturing these cases, we improve our block identification accuracy to 84\%
\item If the next step after a decision point also starts with a conditional, and they have a significant overlap, we can combine the next step as a part of the decision block. For example, if the condition is ``If the slot status is \textit{missing}'', and the next step starts with ``If the slot status is \textit{failed}'', the next step should be a part of the decision block starting at the first condition.
\item About 16 mistakes from 50 were cases where decision blocks end at the sub-list item or paragraph (inside the step) in which the conditional sentence is found. A simple rule to identify these would bump our accuracy further to about 90\%. However, we currently do not keep information about paragraph boundaries or sublists inside list texts, and hence are not able to capture these cases.
\end{itemize}

\subsection{Mapping of Instructions within Decision Blocks}
\begin{figure}
\includegraphics[width=0.9\linewidth]{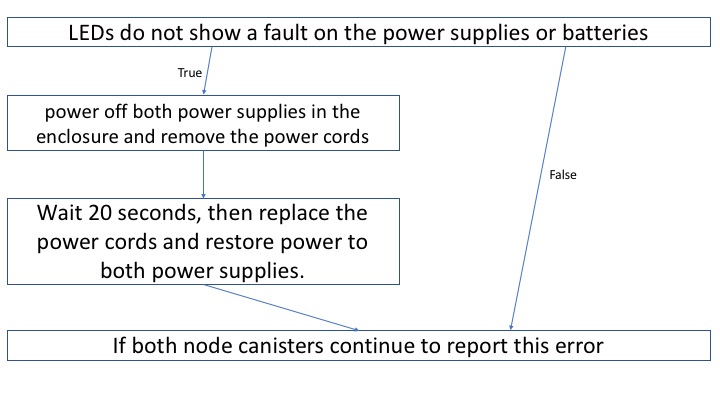}
\caption{Flow of instructions in decision block extracted in Figure~\ref{procexample}(a)}
\label{decision_block_flow_chart}
\end{figure}

\begin{figure}
\includegraphics[width=0.9\linewidth]{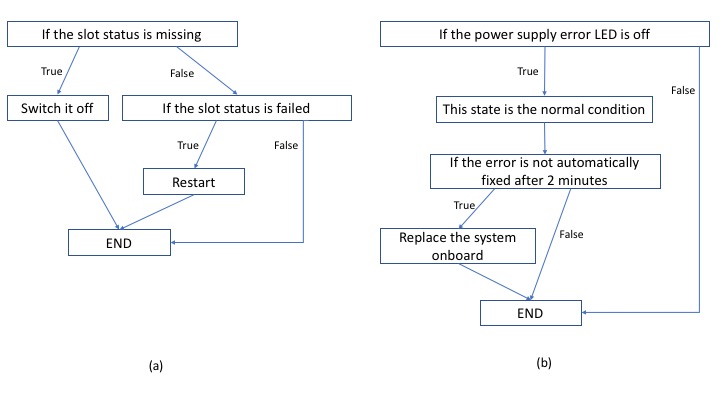}
\caption{Nesting of conditionals inside (a) the false branch, and (b) the true branch}
\label{fig:mapping_example}
\end{figure}
After correctly extracting decision blocks, the next step is to map the instructions present inside the block to the correct path of the split created by the decision point. Figure~\ref{decision_block_flow_chart} depicts a flowchart for the decision block identified in Figure~\ref{procexample}(a). The instructions following the decision point in the block need to be mapped to the ``True'' branch of the flow. From analysis of the data, instructions can be mapped to the true branch of the decision point, till the first instruction with ``else'' or ``otherwise'' is encountered. Encountering other conditional sentences inside a decision block can also potentially change the mapping of the following sentences to the ``False'' branch. However, not all conditional sentences inside a decision block lead to a change in mapping of following instructions. The two possible cases are -
\begin{itemize}
\item If the conditional is parallel to the original condition, the instructions following it need to be mapped to the ``False" branch.
For example, the conditional \textit{If slot status is failed, then restart} indicates a parallel decision for the decision block starting at \textit{If slot status is missing, then switch it off.}
\item The conditional could merely be nested inside the ``True'' branch and not be an alternate path to the original decision point. For example, \textit{If the power supply error LED is off, this state is the normal condition. If the error is not automatically fixed after 2 minutes, replace the system board.} In this example the second condition will only be reached if the first condition is true.
\end{itemize}

The flow of instructions for the two cases is illustrated in Figure~\ref{fig:mapping_example}. To identify if a conditional appearing inside the decision block is nested on the ``True'' or ``False'' branch, we compare the text similarity of the condition to the original decision point. A high text similarity indicates a parallel condition and hence nesting on the ``False'' branch, while a low similarity indicates a nested condition on the ``True'' branch. We find that by thresholding similarity above 70\%, we correctly capture about 85\% of the mappings. However, the absence of a significant amount of data, impedes our effort of analyzing this problem in further detail.

\section{Usecase: Support Chatbot}
Our system for mining procedures is deployed as part of the document ingestion pipeline for a production chatbot for support on IBM DB2, Storwize, and Campaign products. The chatbot, DESIRE \cite{10.1007/978-3-319-68136-8_26}, has an aim of responding to user questions with the most appropriate content found in the collection. It proceeds by creating a knowledge graph from the content consisting of entities like \textit{Component}, \textit{Symptoms}, and \textit{Intents}. It also contains a question understanding module which extracts instances of these entities from the user question and matches them to nodes in the Knowledge Graph and asks back questions to reach leaf nodes (pages in the collection) unambiguously.

\begin{figure*}
\includegraphics[width=\textwidth]{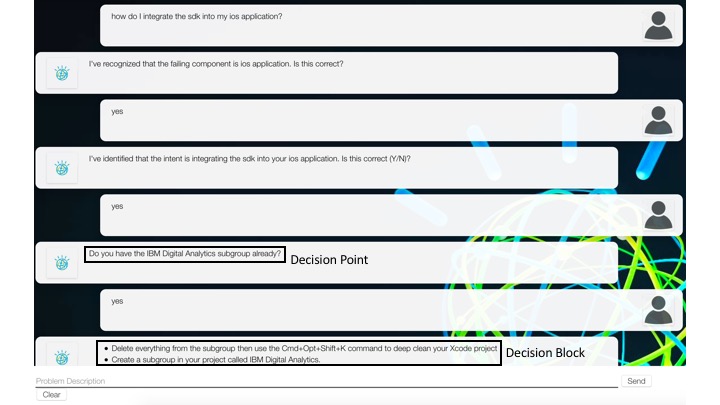}
\caption{Conversation on DESIRE with the procedure for \textit{integrating the SDK into an iOS application, with the decision point and block highlighted}}
\label{DESIRE-example}
\end{figure*}
Procedures identified and extracted from our system are attached to the pages in which they are found during ingestion. When the chatbot reaches such a page containing a procedure, it takes the user through the procedure step-by-step. For steps which have been identified as a decision point the chatbot generates questions from the ``condition'' part. We use a simple template based question generation targeting only Yes/No questions~\cite{heilman2011automatic}. User responses to the questions are mapped to a particular branch within the decision block by identifying the presence or absence of negations in the extracted condition. Figure~\ref{DESIRE-example} shows the example of a conversation in DESIRE with the procedure extracted from our system. Notice the question \textit{Do you have the IBM Digital Analytics subgroup already?} which is created from the conditional sentence \textit{If you already have the IBM Digital Analytics subgroup, delete everything from the subgroup then use the Cmd+Opt+Shift+K command to deep clean your Xcode project}. The following instruction \textit{Create a subgroup in your project called IBM Digital Analytics} is also a part of the decision block, and is displayed together.

\section{Conclusion and Future Work}
We present a system for mining procedures from technical support documents that involves extraction and identification of procedures in text, identification of decision points and decision blocks, and mapping of instructions within the decision block. For procedure identification, we implement a simple baseline, and show how features like the type of list can improve upon the baseline. We also emphasize the extraction of ``imperatives'', and demonstrate their effectiveness at the task, especially in the absence of list type features. For the task of decision block extraction, we show that a simple baseline combined with a handful of rules achieves acceptable performance. Given the sparse data for the task of mapping instructions within a decision block, we are unable to perform detailed analysis. However the few examples underline the difficulty of the task, and its importance in correctly extracting the procedure for downstream applications like a chatbot. We integrate the procedure extraction system into a production chatbot, which improves user experience by leading the user through a procedure versus simply returning urls. We also plan to release all public data, and our annotations used for experiments in this paper to promote further research on the topic.

In this paper we focus mostly on web HTML documents. However, we are also investigating good document conversion tools that can preserve the HTML structure while converting from pdf or word documents. This allows our procedure identification to be generic across types of documents.  
The presence of list structure features like step boundaries helps us get acceptable performance on the task of decision block extraction. In the absence of such features, it still remains a difficult and open problem. Discourse based approaches seem promising for this task. However, we did not find success in applying them on this dataset. Mapping of sentences within a decision block to the correct path of execution at the split created by a decision point also remains open. We plan to collect more data to analyze the problem in detail, as we believe incorrect extraction of such blocks or mapping can adversely affect applications like Robotic Process Automation or chatbot.

\bibliographystyle{ACM-Reference-Format}
\bibliography{sample-bibliography}

\end{document}